\documentclass[10pt,twocolumn,letterpaper]{article}

\usepackage{iccv}
\usepackage{times}
\usepackage{epsfig}
\usepackage{graphicx}
\usepackage{amsmath}
\usepackage{amssymb}
\usepackage[accsupp]{axessibility}
\usepackage{makecell}
\usepackage{multirow}
\usepackage{adjustbox}
\usepackage{hhline}  % 双线表
\usepackage{booktabs}
\usepackage{tablefootnote}
\usepackage{threeparttable}
\usepackage{bm}
\usepackage{hyperref}
\hypersetup{
    colorlinks=true,
    linkcolor=blue,
    filecolor=magenta,      
    urlcolor=cyan,
    pdftitle={Overleaf Example},
    pdfpagemode=FullScreen,
    }

\usepackage{color}

\iccvfinalcopy % *** Uncomment this line for the final submission

\def\iccvPaperID{3124} % *** Enter the ICCV Paper ID here

\ificcvfinal\fi

\begin{document}

%%%%%%%%% TITLE
\title{Multitask AET with Orthogonal Tangent Regularity for Dark Object Detection}

% \author{Ziteng Cui\\
% \and
% Guo-Jun Qi\\

% \and
% Lin Gu\\

% \and
% Shaodi You\\

% \and
% Zenghui Zhang\\

% \and
% Tatsuya Harada\\
\author{Ziteng Cui\textsuperscript{1},
    Guo-Jun Qi\textsuperscript{2},
    Lin Gu\textsuperscript{3,4}\thanks{Corresponding author.}, 
    Shaodi You\textsuperscript{5},
    Zenghui Zhang\textsuperscript{1},
    Tatsuya Harada\textsuperscript{4,3}\\
    \textsuperscript{1}Shanghai Jiao Tong University \textsuperscript{2}, Seattle Research Center, Innopeak Technology\\\textsuperscript{3}RIKEN AIP, \textsuperscript {4} The University of Tokyo,
    \textsuperscript {5} University of Amsterdam \\
    \small{cuiziteng@sjtu.edu.cn, guojunq@gmail.com, lin.gu@riken.jp, s.you@uva.nl, zenghui.zhang@sjtu.edu.cn, harada@mi.t.u-tokyo.ac.jp}
    }

\maketitle
\ificcvfinal\thispagestyle{empty}\fi

%%%%%%%%% ABSTRACT
\begin{abstract}
Dark environment becomes a challenge for computer vision algorithms owing to insufficient  photons and undesirable noise. To enhance object detection in a dark environment, we propose a novel multitask auto encoding transformation  (MAET) model which is able to explore the intrinsic pattern behind illumination translation. In a self-supervision manner, the MAET learns the intrinsic visual structure by encoding and decoding the  realistic illumination-degrading transformation considering the physical noise model and image signal processing (ISP). Based on this representation, we achieve the object detection task by decoding the bounding box coordinates and classes. 
To avoid the over-entanglement of two tasks, our MAET disentangles the object and degrading features by imposing an orthogonal tangent regularity. This forms a parametric manifold along which multitask predictions can be geometrically formulated by maximizing the orthogonality between the tangents along the outputs of respective  tasks.  Our framework can be implemented based on the  mainstream  object detection architecture and directly trained end-to-end using normal target detection datasets, such as VOC and COCO. We have achieved the state-of-the-art performance using synthetic and real-world datasets. Code is available at \href{https://github.com/cuiziteng/ICCV_MAET}{https://github.com/cuiziteng/MAET}.
\end{abstract}

%%%%%%%%% BODY TEXT
\section{Introduction}

\begin{figure}
    \centering
    \includegraphics[width = 8cm, height = 7.5cm]{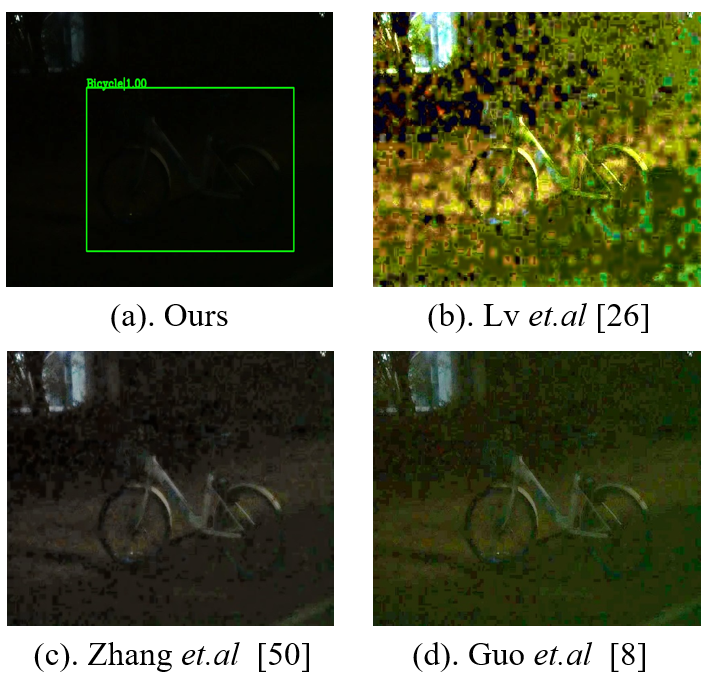}
    \caption{Detection and enhancement results of image taken by Sony DSC-RX100M7 camera at night with 0.1s exposure time and 3200 ISO.   (a) is detection result on the original image by MAET (YOLOv3) and (b), (c), (d) are the enhanced images by Lv \textit{et al.} \cite{Lv2018MBLLEN}, Zhang \textit{et al.} \cite{kind_kill_the_darkness}, Guo \textit{et al.} \cite{zero_dce} respectively, on which YOLOv3 failed to make detection.}
    \label{fig:demo}
\end{figure}

Low-illumination environment poses significant challenges in computer vision. Computational photography community has proposed many human-vision-oriented algorithms to recover normal-lit images ~\cite{LLNet, Chen2018Retinex, Lv2018MBLLEN, lv2021attention, see_in_the_dark, LightenNet, kind_kill_the_darkness, enlightengan, zero_dce}.  Unfortunately, the restored image does not necessarily benefit the high-level visual understanding tasks. As the enhancement/restoration approaches are optimized for human visual perception, they may generate artifacts (see Fig.~\ref{fig:demo} for an example), which are misleading for consequent vision tasks.

Another line of research  focuses on the robustness of specific high-level vision algorithms. They either train models on a large volume of real-world data  ~\cite{NightOwls, LOH201930, UG2} or rely on carefully designed  task-related features ~\cite{traditional_feature_method, VJ_detector_low_light}. 

However, existing methods suffer from two major inconsistencies: \textit{target inconsistency and data inconsistency} (in the existing research).  \textit{Target inconsistency} refers to the fact that most methods focus on their own target, either human vision or machine vision.  Each line follows their routes separately without benefiting each other under a general framework.

In the meantime, \textit{data inconsistency} complicates the assumption that the training data should resemble the one used for evaluation. For example,  pre-trained object detection models are usually trained on clear and normal lit images. To adapt to the poor light condition, they rely on the augmented dark images to fine-tune the models without exploring the intrinsic structure under the illumination variance. Just like \textit{Happy families are all alike; every unhappy family is unhappy in its own way}, even with the existing datasets ~\cite{voc,coco_dataset,krizhevsky2017imagenet}, the varied distribution of real-world conditions can hardly be covered by the training set.

\begin{figure}
    \centering
    \includegraphics[width = 8.6cm, height = 5.4cm]{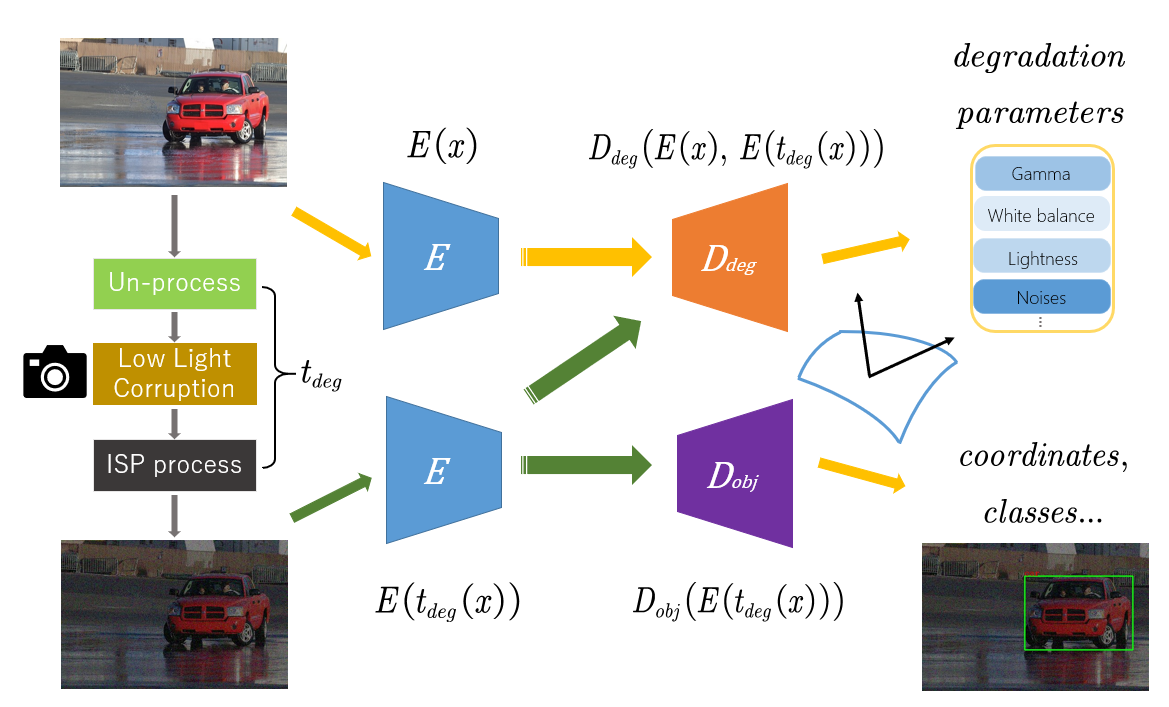}
    \caption{Structure of the  multitask autoencoding transformations (MAET) framework.}
    \label{fig:illu_maet}
\end{figure}

Here, we aim to bridge above two gaps under a unified framework. As illustrated in Fig.~\ref{fig:illu_maet}, the normal lit image can be parametrically  transformed ($t_{deg}$) into their degraded low-illumination counterparts. Based on this transformation, we propose a novel multitask autoencoding transformation (MAET) to extract the transformation equivariant convolutional features for object detection in dark images. We train the MAET based on two tasks: (1) to learn the intrinsic representation by decoding the low-illumination-degrading transformation based on unlabeled data and (2) to decode object position and categories based on labeled data. As shown in Fig.~\ref{fig:illu_maet}, we train our MAET to encode the pair of normal lit and low-light images with  siamese encoder $E$  and decode its degrading parameters, such as noise level, gamma correction, and white balance gains, using decoder $D_{deg}$. This allows our model to capture the intrinsic visual structure that is equivariant to illumination variance. Compared with ~\cite{LLNet, shen2017msrnetlowlight, LightenNet, Lv2018MBLLEN, yolo-in-the-dark}, who conducted over-simplified synthesis, we design our degradation model considering the physical noise model of sensors and  image signal processing (ISP). Then, we perform the object detection task  by decoding the bounding box coordinates and classes with the decoder $D_{obj}$ based on the representation encoded by $E$ (Fig.~\ref{fig:illu_maet}).

Although MAET regularizes network training by predicting low-light degrading parameters, the joint training of object detection and transformation decoding are over-entangled through a shared backbone network. While this improves the detection of dark objects using MAET regularity, it may also risk overfitting the object-level representation into self-supervisory imaging signals.  To this end, we propose to disentangle the object detection and transformation decoding tasks by imposing an orthogonal tangent regularity.  It assumes that the multivariate outputs of above two tasks form a parametric manifold, and  disentangling the multitask outputs along the manifold can be geometrically formulated by maximizing the orthogonality among the tangents along the output of different  tasks. The framework can be directly trained end-to-end using standard target detection datasets, such as COCO \cite{coco_dataset} and VOC \cite{voc}, and make it detect low-light images. Although we consider YOLOv3 ~\cite{yolov3} for illustration, the proposed MAET is a general framework that can be easily applied to other mainstream object detectors, \textit{e.g.}, \cite{fasterrcnn,retinanet,centernet}.

Our contributions to this study are as follows:
\begin{itemize}
     \item  By exploring physical noise models of sensors and the ISP pipeline, we leverage a novel MAET framework to encode the intrinsic structure, which can decode low-light-degrading transformation. Then, we perform the object detection by decoding bounding box coordinates and categories based on this robust representation. Our MAET framework is compatible  with  mainstream object architectures. 
   \item  Moreover, we present the disentangling of multitask outputs to avoid the overfitting of the learned object-detection features into the self-supervisory degrading parameters. This can be naturally performed from a geometric perspective by maximizing the orthogonality along the tangents corresponding to the output of different tasks.
    \item Based on comprehensive evaluation and compared with other methods, our method shows superior performance pertaining to low-light object detection tasks. 
\end{itemize}

\section{Related Work}

\subsection{Low Illumination Datasets}

Several datasets have been proposed for the low-light object detection task: Neumann \textit{et al.} ~\cite{NightOwls} proposed NightOwls dataset for pedestrians detection in the night. Nada \textit{et al.} ~\cite{Unconstrained_Face_Detection} collected an unconstrained face detection dataset (UFDD) considering various adverse conditions, such as rain, snow, haze, and low illumination. In recent times, the $\rm UG^2+$ challenge ~\cite{UG2} has included several tracks for vision tasks under different poor visibility environments. Among them, DARK FACE dataset  with 10,000 images (includes 6,000 labeled and 4,000 unlabeled images).  For the multi-class dark object detection task, Loh \textit{et al.} ~\cite{LOH201930} proposed exclusively dark (ExDark) dataset, which includes 7363 images with 12 object categories.

%------------------------------------------------------------------------
\subsection{Low-Light Vision}

\subsubsection{Enhancement and Restoration Methods}

Low-light vision tasks focus on the human visual experience by restoring details and correcting the color shift. Early attempts are either Retinex theory based approaches ~\cite{retinex, multiscale_retinex, LIME_Illumination_Map_Estimation} or histogram equalization (HE) based approaches ~\cite{Global_HE, local_HE}. Nowadays, with the development of deep learning, CNN based methods ~\cite{LLNet, Chen2018Retinex, Lv2018MBLLEN, lv2021attention, LightenNet, kind_kill_the_darkness, zero_dce} and GAN based methods ~\cite{low_light_gan, enlightengan} have achieved a significant improvement in this task. Like Wei \textit{et al.} ~\cite{Chen2018Retinex} combined the Retinex theory ~\cite{retinex} with deep network for low-light image enhancement. Jiang \textit{et al.}  ~\cite{enlightengan} used an unsupervised GAN to solve this problem. Very recently, Guo \textit{et al.} ~\cite{zero_dce} proposed a self-supervised method, which could learn without normal light images.

\subsubsection{High-Level Task}

To adopt the high-level task for a dark environment, a straightforward strategy is casting the aforementioned enhancement methods as a post-processing step ~\cite{zheng_2020_ECCV, zero_dce}. Other ones rely on augmented real-world data~\cite{NightOwls, LOH201930, UG2, low_light_saliency} or some oversimplified synthetic data  \cite{optical_flow_in_the_dark, yolo-in-the-dark}.%  Other researchers enhance their visual understanding methods with manually crafted task-related features 
% e.g., parse representation prior ~\cite{traditional_feature_method}.
 Recent real noisy image benchmarks ~\cite{ANAYA2018NENOIR,Plotz17CVPR} show that sometimes hand-crafted algorithms may even outperform deep learning models. To combine the strength of computational photography, we develop a framework with transformation-equivariant representation learning. 

\subsection{Transformation-Equivariant Representation Learning}

Several self-supervised representation learning methods have been proposed to learn image features either through solving Jigsaw Puzzles  ~\cite{jigsaw_selfsupervised} or impainting the missing region of an image ~\cite{pathak2016context}.  Recently, a series of  auto-encoding transformations (AETs), such as  AET ~\cite{zhang2019aet}, AVT ~\cite{qi2019avt}, EnAET ~\cite{wang2019enaet}, have demonstrated state-of-the-art performances for several self-supervised tasks. As the AET is flexible and not restricted to any specific convolutional structure, we extend it to our multitask AET for object detection in dark images.

\section{Multitask Autoencoding Transformation (MAET)}

In this section, we first briefly introduce auto-encoding transformation (AET) ~\cite{zhang2019aet}, based on which we propose multitask AET (MAET). Then, we discuss the ISP pipeline in camera to design the degrading transformations to be leveraged by our MAET. Finally, we explain the MAET architecture and training and testing details.

\subsection{Background: From AET to MAET}
\label{aet_model}

AET ~\cite{zhang2019aet} learns representative latent features that decode or recover the parameterized transformation from the original image ($x$) and the transformed counterpart ($t(x)$) based on transformation $t$:
\begin{equation}
    x \underrightarrow{\ \ \  \mathcal{T}\ \ \  } t(x) \\. 
\label{eq:aet_1}
\end{equation}

The AET comprises a siamese representation encoder ($E$) and a transformation decoder ($D$). The encoder $E$ extracts features from $x$ and its transformation $t(x)$, which should capture intrinsic visual structures to explain the transformation $t$ (\textit{e.g.}, the low-illumination degrading transformations in the next section). Then, the decoder $D$ uses the encoded $E(x)$ and $E(t(x))$ to decode the estimation $\hat{t}$ for $t$:
\begin{equation}
   \hat{t}= D_\phi[E(x),E(t(x))].
\label{eq:aet_decoder}
\end{equation}

The AET, specifically the representation encoder $E$ and transformation decoder $D$, can be trained by minimizing the deviation loss $\ell$ of the original transform $t$ and the predicted result $\hat{t}$:
\begin{equation}\label{eq:loss_aet}
    \mathcal L_{aet} \triangleq \sum_k \ell_k(\hat{t_k}, t_k),
\end{equation}
where  $\ell_k$ denotes type $k$ transformation loss computed using the  mean-squared error (MSE) loss between the predicted transformation $\hat{t_k}$  and ground truth transformation $t_k$.

\subsection{Multi-Task AET with Orthogonal Regularity}
In this study, we further extend the AET to the MAET by simultaneously solving multiple tasks. 
As illustrated in Fig.~\ref{fig:illu_maet}, the proposed MAET model consists of two parts: a representation encoder ($E$) and multi-task decoders. For the task of illumination-degrading transformation $t_{deg}$, we use decoder $D_{deg}$ to decode the degrading parameters. The task of object detection is realised by the decoder $D_{obj}$ to predict the bounding box location and object categories directly from illumination-degenerated images. Although the two tasks are correlated, their outputs reflect very different aspects of input images: the illumination conditions for $D_{deg}$ and the object locations and categories for $D_{obj}$. This suggests that an orthogonal regularity can be imposed to decouple the unnecessary interdependence between the outputs of different tasks.

To this end, the orthogonal objective of the proposed MAET is to minimize the absolute value of cosine similarity below:

\begin{equation}\label{eq:loss_ort}
  \mathcal L_{ort} \triangleq \sum_{k,l} |\cos \theta_{k,l}| = \sum_{k,l} \frac{|\left[ \frac{\partial E}{\partial D^k_{deg}} \right] ^T\cdot \left[ \frac{\partial E}{\partial D^l_{obj}} \right]|}{\lVert \left. \frac{\partial E}{\partial D^k_{deg}} \rVert \cdot \lVert \frac{\partial E}{\partial D^l_{obj}} \rVert \right.},
\end{equation}
where $\frac{\partial E}{\partial D^k_{deg}}$ and $\frac{\partial E}{\partial D^l_{obj}}$ are the tangents of the representation manifold formed by the encoder $E$ along the $k$th and $l$th output coordinates of the illumination-degrading transformation and object detection tasks, respectively.  In other words, these two tangents depict the directions along which the representation moves with the change of the decoder outputs $D^k_{deg}$ and $D^l_{obj}$, respectively. 

%$D^k_{aet}$ and $D^l_{obj}$ \footnote{Here, for simplicity, $D_{aet}$ and $D_{obj}$ each denotes one of AET and object prediction coordinates. This tangent orthogonality regularity should add up all such $|\cos\theta|$ over all pairs of coordinates between two tasks. }, respectively. 

Minimizing the absolute value of the cosine similarity will push the two tangents as orthogonal as possible. Based on the geometric point of view, this will disentangle the two tasks so that the change of the predicted coordinates for one task will have a minimal impact on the coordinates for the other task. In Sections~\ref{sec:degrading}, we will discuss the details about how to define the low-illumination-degrading transformation. The idea of imposing orthogonality between tasks was explored in literature~\cite{suteu2019regularizing,xiumei,yu2020gradient}. However, here we implement it in the context of AET, where the orthogonal directions are defined in terms of decoder tangents along the encoder-induced manifold, which differs from the previous works.

Therefore, the total loss for our low-light object detection consists of three parts: degradation transformation loss $\mathcal L_{deg}$, object detection loss $\mathcal L_{obj}$ and orthogonal regularity loss $\mathcal L_{ort}$ (cf. Eq.~(\ref{eq:loss_ort})), the total loss used for training can be represented as
\begin{equation}\label{loss_function}
   \mathcal L_{total} = \mathcal L_{ort} + \omega_1 \cdot \mathcal L_{obj} + \omega_2 \cdot \mathcal L_{deg} .
\end{equation}
The  object detection loss $\mathcal L_{obj}$  is specific for different object detectors ~\cite{fasterrcnn,centernet,yolov3}. In this experiments, $\mathcal L_{obj}$ is the loss function of YOLOv3 ~\cite{yolov3}, which includes location loss, classification loss and confidence loss. The degradation transformation loss $\mathcal L_{deg}$ is the AET loss (cf. Eq.~(\ref{eq:loss_aet})) with the low-illumination degrading transformation $t_{deg}$, and $\omega_1$  and $\omega_2$ are the fixed balancing hyper-parameters.

%For the low-light object detection task, $\mathcal L_{total}$ can be represented as
% \begin{equation}
% \label{loss_function}
%   \mathcal L_{total} = \mathcal L_{ort} + \omega_1 \cdot \mathcal L_{aet \ obj} + \omega_2 \cdot \mathcal L_{aet \ deg}  
% \end{equation}

\subsection{Low-Illumination Degrading Transformations}\label{sec:degrading}
%Here, we will elaborate on the notion of transformations in this section. 

Given a normal lit noise-free image $x$, we aim to design a low-illumination-degrading transformation $t_{deg}$ to transform $x$ into a dark image $t_{deg}(x)$ that matches the real photo captured under low-light conditions, \textit{i.e.}, by turning off the light. Most of existing methods conduct an over-simplified synthesis, \textit{e.g.}, invert gamma correction (sometimes with  additive mixed Gaussian noise) \cite{LLNet, Lv2018MBLLEN, optical_flow_in_the_dark} or retinex theory based synthesis method ~\cite{LightenNet}.  The ignorance of the physics of sensors and on-chip image signal processing (ISP) makes these methods generalise poorly to real-world dark images. Here, we first  systematically describe the ISP pipeline between the sensor measurement system and the final photo. Based on this pipeline, we parametrically model the low light-degrading transformation  $t_{deg}$. 

%Then we explain the object detection as transformation of coordinates. 

%\subsubsection{Low–Illumination-Degrading Transformations}

\begin{figure*}
    \centering
    \includegraphics[width = 14.5cm, height = 8cm]{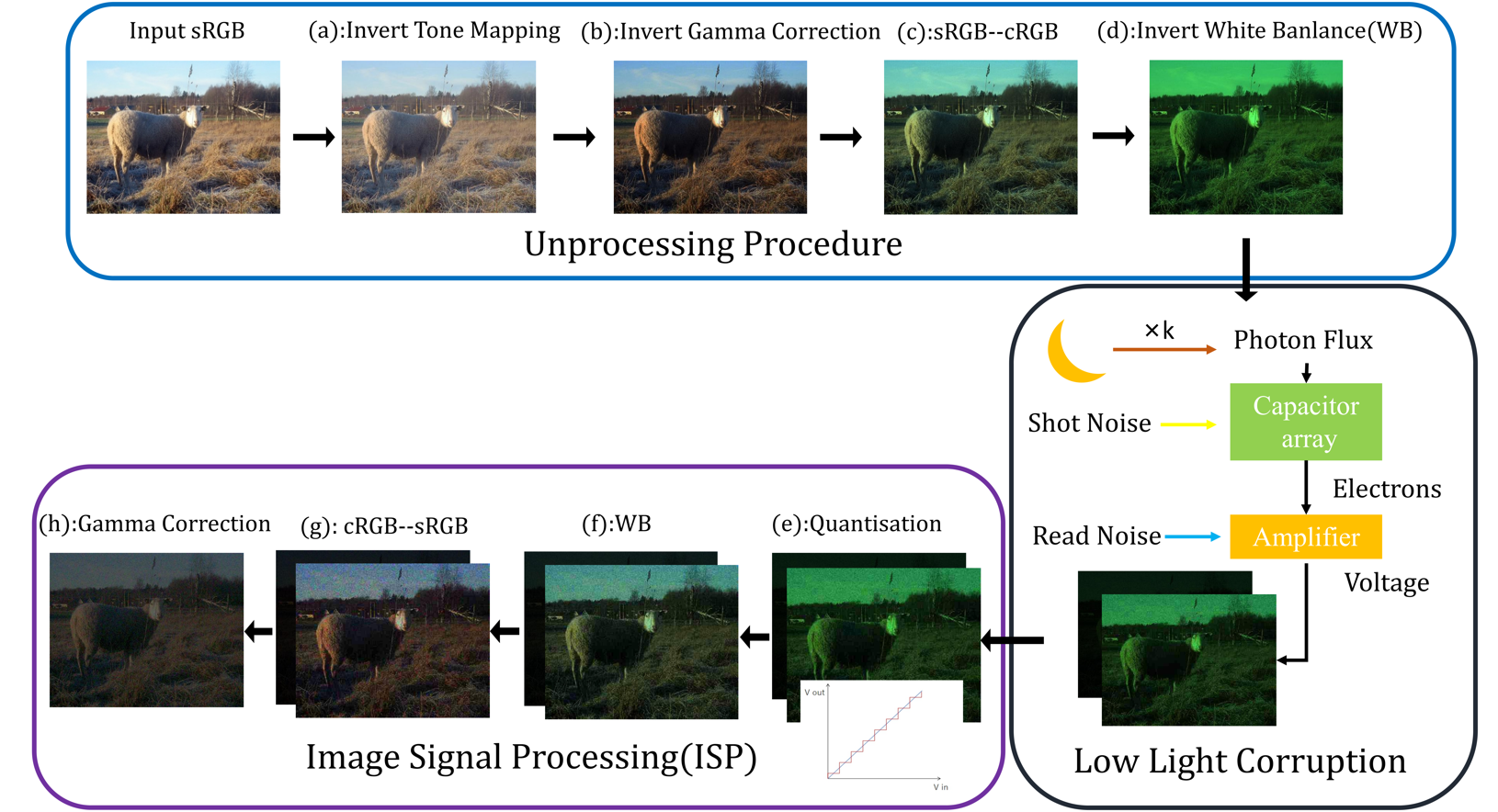}
    \caption{General view of the low-illumination-degrading pipeline, an sRGB "sheep" image from \textit{PASCAL VOC2007  dataset} ~\cite{voc} passed by unprocessing procedure, low-light corruption, and image signal processing(ISP) process to get the final degraded low-light counterpart}
    \label{fig:pipeline} 
\end{figure*}

%Our photosensor model is primarily based upon the CCD sensor.
\subsubsection{Image Signal Processing (ISP) Pipeline}

The camera is designed to render the photo to be as pleasant and accurate as possible based on the perspective of a human eye. For this reason, the RAW data captured by the camera sensor requires ISP (several steps) before becoming the final photo. 
Much research has been done to simulate this ISP process ~\cite{ISP_2005_magazine, ISP_flexisp, ISP_L3method, isp_icip_2018, mobile_camera_imaging2}. For example, Karaimer and Brown ~\cite{ISP_ECCV_2016_brown} step-by-step detailed the ISP process and showed its high potential pertaining to computer vision. 

We adopt a simplified ISP and its unprocessing procedure from \cite{brooks2019unprocessing} (Fig.~\ref{fig:pipeline}). Particularly, we ignore several steps including the demosaicing process ~\cite{demosaic}.  Although these processes are important for precise ISP algorithm, most images on the Internet are of various sources and do not follow the perfect ISP procedure. We ignore these steps for the trade-off between precision and generability. We have made a detailed analysis of the demosaicing's influence in supplementary materials Appendix.B.2. Next, we introduce our ISP process in detail.

\textbf{Quantization} is the analog voltage signal step that quantizes the analog measurement  $x$ into discrete codes $y_{quan}$ using an analog-to-digital converter (ADC). The quantization step maps a range of analog voltages to a single value and generates a uniformly distributed quantization noise. To simulate the quantization step, the quantization noise $x_{quan}$ related to $B$ bits has been added. In our degrading model, $B$ is randomly chosen from 12, 14, and 16 bits.

\begin{equation}
\begin{aligned}
    & x_{quan} \sim U(-\frac{1}{2B}, \frac{1}{2B}) \\
    & y_{quan} = x + x_{quan}.
\end{aligned}    
\end{equation}

\textbf{White Balance}  simulates  the color constancy of human vision system (HVS) to map “white” colors with the white object \cite{ISP_2005_magazine}. The captured image is the product of the color of light and material reflectance. The white-balance step in the camera pipeline estimates and adjusts  the red channel gain $g_r$ and blue channel gain  $ g_b$  to make image appearing to be lit under "neutral" illumination. 
%To avoid the saturation issue, rather than directly apply the reciprocal of red and blue gain $1/g$, we follow \cite{brooks2019unprocessing} to conduct white balance following the below equation.
\begin{equation}
\left[ \begin{array}{c}
	y_r\\
	y_g\\
	y_b\\
\end{array} \right] =\left[ \begin{matrix}
	g_r&		0&		0\\
	0&		1&		0\\
	0&		0&		g_b\\
\end{matrix} \right] \cdot \left[ \begin{array}{c}
	x_r\\
	x_g\\
	x_b\\
\end{array} \right]
\end{equation}

Based on ~\cite{Plotz17CVPR, brooks2019unprocessing}, $g_r$ is randomly chosen from $(1.9, 2.4)$, and $g_b$ is randomly chosen from $(1.5, 1.9)$; both follow an uniform distribution and are independent of each other. The inverse process considers the reciprocal of the red and blue gains $1/g$.

\textbf{Color Space Transformation} converts the white-balanced signal from camera internal color space $cRGB$ to $sRGB$ color space. This step is essential in ISP pipeline as camera color space are not identical to the sRGB space ~\cite{ISP_2005_magazine, ISP_ECCV_2016_brown}. The converted signal $y_{sRGB}$ can be obtained with a  $3\times 3$ color correction matrix (CCM) $M_{ccm}$:
\begin{equation}
y_{sRGB} = M_{ccm} \cdot y_{cRGB},
\end{equation}
the inversion of this process is:
\begin{equation}
y_{cRGB} = M_{ccm}^{-1} \cdot y_{sRGB}.
\end{equation}

\textbf{Gamma Correction} has also been widely used in the ISP pipeline for the non-linearity  of humans perception on dark areas ~\cite{poynton2003digital}. Here we use the standard gamma curve ~\cite{Plotz17CVPR} as:
\begin{equation}
    y_{gamma} = max(x, \epsilon)^{\frac{1}{\gamma}}
\label{eq:gamma}
\end{equation}
and its' invert process is:
\begin{equation}
    y_{invert \ gamma} = max(x, \epsilon )^{\gamma}.
\label{eq:gamma}
\end{equation}

The gamma curve parameter $\gamma$ could be randomly sampled from an uniform distribution $\gamma \sim U(2,\;3.5)$ and $\epsilon$ is a very small value ($\epsilon = 1e^{-5}$) to prevent numerical instability during training. 

\begin{figure}
    \centering
    \includegraphics[width = 8.5cm]{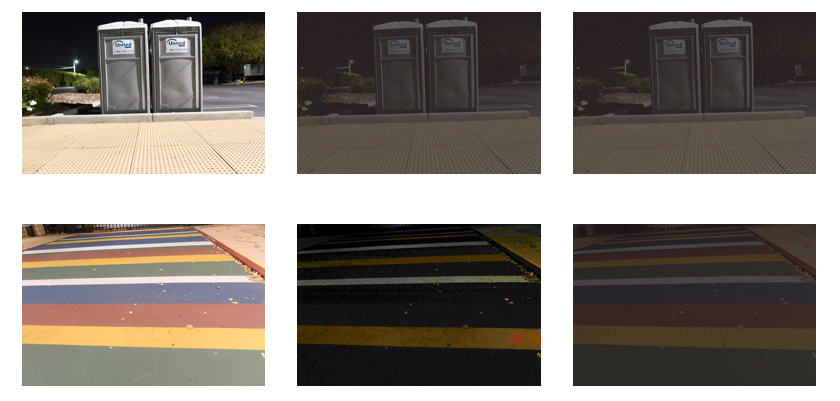}
    \caption{Examples of our degrading transformation on SID dataset \cite{see_in_the_dark}. The long-exposure RAW images and their ground truth short-exposure RAW images are transformed into the sRGB format using $Adobe \ Lightroom$, separately shown in the first and second columns. The third column shows the images generated from our pipeline.}
    \label{fig:syn_example}
\end{figure}

\textbf{Tone Mapping} aims to match the “characteristic curve” of film. For the sake of computational complexity, we perform a  "smoothstep" curve ~\cite{brooks2019unprocessing} as 
\begin{equation}
    y_{tone} = 3x^2 - 2x^3
\end{equation}
and we could also perform the inverse with:
\begin{equation}
   y_{invert\ tone} = \frac{1}{2} - sin(\frac{sin^{-1}(1-2x)}{3}).
\end{equation}
 
\subsubsection{Degrading Transformation Model}

After defining each step of our ISP pipeline, we can present our low-illumination-degradation transform $t_{deg}$ that synthesizes realistic dark light image $t_{deg}(x)$ based on its normal light counterparts $x$. At first, as shown in Fig.~\ref{fig:pipeline}, we have to use an inverse processing procedure~\cite{brooks2019unprocessing} to transform the normal lit image $x$ into sensor measurement or RAW data. Then, we linearly attenuate the RAW image and corrupt it with shot and read noise. Finally, we continue applying the pipeline to turn the low-lit sensor measurement to the photo $t_{deg}(x)$. 

\textbf{Unprocessing Procedure}:
Based on ~\cite{brooks2019unprocessing}, the unprocessing part aims to translate the input $sRGB$ images into their RAW format counterparts, which are linearly proportional to the captured photons. As shown in Fig.~\ref{fig:pipeline}, we unprocess the input images by (a) invert tone mapping, (b) invert gamma correction, (c) transformation of image from $sRGB$ space to $cRGB$ space, and (d) invert white balancing, here we call (a), (b), (c), (d) together as $t_{unprocess}$. Based on these parts, we synthesize realistic RAW format images, and the resulting synthetic RAW image is used for low-light corruption process.

\textbf{Low Light Corruption}:

When light photons are projected through a lens on a capacitor cluster, considering the same exposure time, aperture size, and automatic gain control, each capacitor develops an electric charge corresponding to the lux of illumination of the environment.

Shot noise is a type of noise generated by the random arrival of photons in a camera, which is a fundamental limitation. As the time of photon arrival is governed by Poisson statistics, uncertainty in the number of photons collected during a given period is  $\delta_{s} = \sqrt{S}$, where $\delta_{s}$ is the shot noise and $S$ is the signal of the sensor.
%, they both expressed in electrons.

Read noise occurs during the charge  conversion of electrons into voltage in the output amplifier, which can be approximated using a Gaussian random variable with zero mean and fixed variance.

Shot and read noises are common in a camera imaging system; thus, we model the  noisy measurement $x_{noise}$ ~\cite{Noise_model} on the sensor: 

\begin{equation}
\begin{aligned}
    &x_{noise} \sim N(\mu = kx, \sigma^{2} = \delta_{r}^2 + \delta_{s}kx)\\
    &y_{noise} = kx + x_{noise},
\end{aligned}
\end{equation}

where the true intensity of each pixel $x$ from the unprocessing procedure. We linearly attenuate it with parameter $k$. To simulate different lighting conditions, the parameter of light intensity $k$ is randomly chosen from a truncated Gaussian distribution, in range of $(0.01, 1.0)$, with mean $0.1$ and variance $0.08$. The parameter range of $\delta_{r}$ and $\delta_{s}$  follows \cite{Plotz17CVPR}, which is shown in Table \ref{tab:trans_para}.

\begin{figure}
    \centering
    \includegraphics[width = 8.5cm]{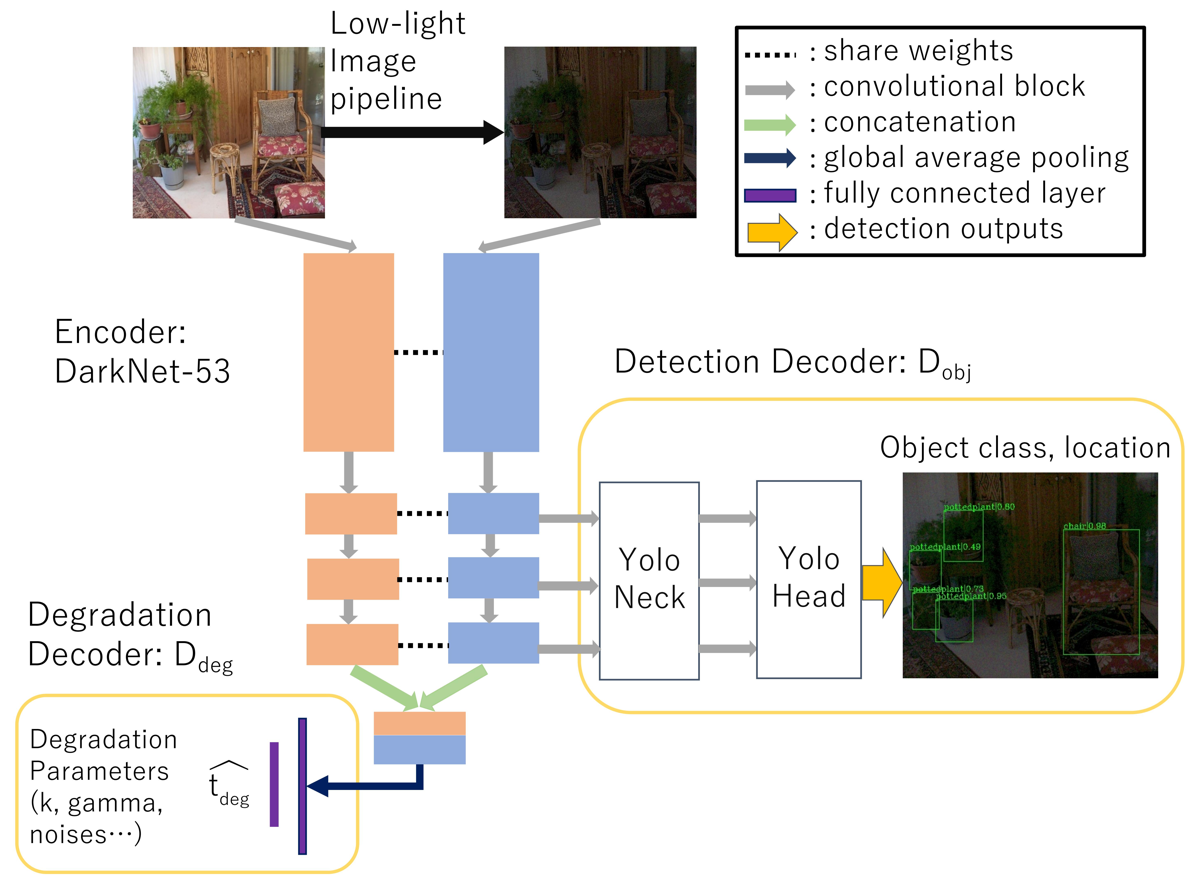}
    \caption{Architecture of the proposed MAET model base on YOLOv3 framework.}
    \label{fig:yolomaet}
\end{figure}

%, $\delta_{r}$ and $\delta_{s}$ are the noise parameters can vary across images as sensor gain(ISO) changes.

\begin{table*}
  \centering
  \begin{adjustbox}{max width= 1 \linewidth}
  \begin{tabular}{c c c c}
  \toprule[1.5pt]
  \toprule[1.5pt]
  Step & Transformation & Range & Parameter(s) to Learn\\
  \midrule[1.2pt]
  \midrule[1.2pt]
  Light Intensity & $f(x)=k \cdot x$ &\makecell[c]{$k\sim N(\mu = 0.1, \sigma = 0.08)$ \\ $0.01 \leq k \leq 1.0 $} & $k$ \\\specialrule{0.05em}{3pt}{3pt}
  
  \makecell[c]{Shot Noise and \\Read Noise} &$f(x)=x+N(\mu = x, \sigma^{2} = \delta_{r}^2 + \delta_{s}x)$ & \makecell[c]{ $log\delta_{s} \sim U(-4, -2)$\\$\frac{log\delta_{r}^2}{log\delta_{s}} \sim N(\mu = 2.18 log\delta_{s} + 0.12, \sigma = 0.26)$} & -\\\specialrule{0.05em}{5pt}{5pt}

  Quantization & \makecell[c]{$f(x) = x + U(-\frac{1}{2B}, \frac{1}{2B})$}& $B \in [12, 14, 16]$ & $\frac{1}{B}$\\\specialrule{0.05em}{3pt}{3pt}

  \makecell[c]{White Balance} & $f(x) = \begin{bmatrix}
   g_r & 0 & 0 \\
   0 & 1 & 0 \\
   0 & 0 & g_b
  \end{bmatrix} \cdot x$& \makecell[c]{$g_r \sim U(1.9, 2.4)$\\$g_b \sim U(1.5, 1.9)$} & $\frac{1}{g_r}, \frac{1}{g_b}$\\\specialrule{0.05em}{5pt}{5pt}

  \makecell[c]{Color Correction \\$cRGB \rightarrow uRGB \rightarrow sRGB$} & \makecell[c]{$f(x) = M_{cu} \cdot M_{us} \cdot x$} & \makecell[c]{Mixture of four color correction \\ matrices (CCMs): Sony A7R, Olympus E-M10, \\Sony RX100 IV, Huawei Nexus 6P in \cite{brooks2019unprocessing}}
  & -
  \\\specialrule{0.05em}{3pt}{3pt}
  
  Tone Mapping & $f(x)=3x^2 - 2x^3$ & - & -\\\specialrule{0.05em}{2pt}{2pt} 

  Gamma Correction & $f(x)=max(x, \epsilon)^{\frac{1}{\gamma}},\ \epsilon = 1e^{-5}$ & $\gamma \sim U(2,\;3.5)$ & $\frac{1}{\gamma}$\\
  \bottomrule[1.5pt]
  \bottomrule[1.5pt]
  \end{tabular}
  \end{adjustbox}
  \caption{Details of low-illumination-degrading transformation parameters, the first column denotes the names of the transformations, the second column denotes the transformation process, the third column is the parameter range, and the last line denotes the parameters to be predicted in our MAET model's degrading transformation decoder.} \label{tab:trans_para}

\end{table*}
% \footnotetext[1]{Four types CCM matrices chosen from Sony A7R, Olympus E-M10, Sony RX100 IV, Huawei Nexus 6P in \cite{8099777}}

\textbf{ISP Pipeline}:
RAW images often pass through a series of transformations, before we see it in the RGB format; therefore, we apply RAW image processing after the low-light corruption process. Based on ~\cite{brooks2019unprocessing}, our transformations are in the following order: (e) add quantization noise (f) white balancing, (g) from $cRGB$ to $sRGB$, and (h) gamma correction,  we call (f), (g), (h) together as $t_{ISP}$.

Finally, we can obtain the degraded low-light images $t_{deg}(x)$ from the noise-free $x$, as it shown in Eq.~\ref{eq:whole_pipeline}. Some examples of the original images, generated images, and ground truth are shown in Fig.~\ref{fig:syn_example}. We summarise the  parameters and their ranges involved in $t_{deg}$ (Table \ref{tab:trans_para}):
\begin{equation}
    t_{deg}(x) = t_{ISP}(k \cdot t_{unprocess}(x) + x_{noise} + x_{quan}).
\label{eq:whole_pipeline}
\end{equation}

%------------------------------------------------------------------------
\subsection{Architecture}

The architecture of the proposed MAET is shown in Fig.~\ref{fig:yolomaet}. Our network comprises representation encoder $E$ and  decoder $D$. For illustration, we implement the MAET based on the architecture of YOLOv3 \cite{yolov3}. Moreover, this can be replaced by other mainstream detection frameworks, \textit{e.g.} , \cite{fasterrcnn}, \cite{retinanet}, \cite{centernet}.

$E$ adopts a siamese structure with shared weights. During the training process, the normal lit image $x$ is fed into the left path of $E$ (denoted in orange), while its  degraded counterparts $t_{deg}(x)$ go through the right path or dark path (denoted in blue). Here, the encoder adopts DarkNet-53 network \cite{yolov3} as the backbone.

As we solve two tasks, degrading transformation decoding and object detection tasks, decoder $D$ can be divided into degrading transformation decoder $D_{deg}$ and object detection decoder $D_{obj}$. The former  focuses on decoding the parameters of low-light-degrading transformation ($t_{deg}$). The latter decodes the target information, \textit{i.e.}, target class and location. As shown in Fig.~\ref{fig:yolomaet}, the encoded latent features $E(x)$ and $E(t_{deg}(x))$ are concatenated together and passed to decoder $D_{deg}$ to estimate the corresponding degrading transformation $t_{deg}$. This self-supervision training helps the MAET learn the intrinsic visual structure under various illumination degrading transformations with unlabeled data. The object detection decoder  $D_{obj}$ only decodes the representation $E(t_{deg}(x))$ from the dark path (denoted in blue) to predict the parameters of object detection. In the testing time, we directly feed the low-light images to the dark path of the MAET encoder to decode the detection results: target categories and locations. 

\section{Experiments}
\label{sec:Experiments}

\begin{table*}[!htb]
\begin{adjustbox}{max width= 1 \linewidth}
\begin{tabular}{|c|c|c|c|ccccccc|}
\hline
                      & training set                & testing set          & pre-process        & \multicolumn{1}{c}{$\rm VOC (AP_{50})$} & \multicolumn{1}{c}{$\rm COCO (AP)$} & \multicolumn{1}{c}{$\rm COCO (AP_{50})$} & \multicolumn{1}{c}{$\rm COCO (AP_{75})$} & \multicolumn{1}{c}{$\rm COCO (AP_S)$} & \multicolumn{1}{c}{$\rm COCO (AP_M)$} & {$\rm COCO (AP_L)$} \\ \hline
\multirow{5}{*}{YOLO} & \multirow{4}{*}{normal}     & normal               & -                  & 0.802                           & 0.335                         & 0.573                           & 0.352                           & 0.195                          & 0.364                          & 0.436     \\ \cline{3-11} 
                      &                             & \multirow{6}{*}{low} & MBLLEN             & 0.712                           & 0.239                         & 0.411                           & 0.243                           & 0.115                          & 0.258                          & 0.342     \\ \cline{4-11} 
                      &                             &                      & KIND               & 0.729                           & 0.254                         & 0.437                           & 0.255                           & 0.138                          & 0.293                          & 0.365     \\ \cline{4-11} 
                      &                             &                      & Zero-DCE           & 0.717                           & 0.250                         & 0.422                           & 0.243                           & 0.129                          & 0.302                          & 0.358     \\ \cline{2-2} \cline{4-11} 
                      & low                         &                      & \multirow{3}{*}{-} & 0.764                           & 0.318                         & 0.522                           & 0.309                           & 0.162                          & 0.344                          & 0.405     \\ \cline{1-2} \cline{5-11} 
\textbf{MAET (w/o ort)}        & \multirow{2}{*}{low+normal} &                      &                    & 0.770                           & 0.321                         & 0.534                           & 0.331                           & 0.163                          & 0.355                          & 0.401     \\ \cline{1-1} \cline{5-11} 
\textbf{MAET (w ort)}          &                             &                      &                    & \textbf{0.788}                           & \textbf{0.330}                         & \textbf{0.569}                           & \textbf{0.341}                           & \textbf{0.189}                          & \textbf{0.362}                          & \textbf{0.421}     \\ \hline
\end{tabular}
\end{adjustbox}
\caption{The experimental results on VOC ~\cite{voc} dataset and COCO ~\cite{coco_dataset} dataset.}\label{tab:VOCCOCO}
\end{table*}

\begin{table*}
\centering
\begin{adjustbox}{max width= 1 \linewidth}
\begin{tabular}{|c|cccccccccccc|c|}
\hline
                                         & Bicycle        & Boat           & Bottle & Bus            & Car            & Cat            & Chair          & Cup            & Dog            & Motorbike      & People         & Table          & \multicolumn{1}{l|}{\textbf{Total}} \\ \hline
YOLO (N)                                  & 0.718          & 0.645          & 0.639  & 0.816          & 0.768          & 0.554          & 0.497          & 0.568          & 0.638          & 0.618          & 0.657          & 0.405          & 0.627                                                             \\ \hline
MBLLEN \cite{Lv2018MBLLEN} + YOLO (N)                         & 0.732          & 0.644          & 0.672  & 0.892          & 0.770          & 0.607          & 0.571          & 0.661          & 0.697          & 0.634          & 0.697          & 0.439          & 0.668                               \\ \hline
KIND \cite{kind_kill_the_darkness} + YOLO (N)                           & 0.734          & 0.681          & 0.655  & 0.862          & 0.783          & 0.630          & 0.569          & 0.627          & 0.682          & 0.671          & 0.696          & 0.482          & 0.673                               \\ \hline
\multicolumn{1}{|l|}{Zero-DCE \cite{zero_dce} + YOLO (N)} & 0.795 & 0.713         & 0.704  & 0.890      & 0.807          & 0.684          & \textbf{0.657} & 0.686          & \textbf{0.754} & 0.672          & 0.762          & 0.511          & 0.720                               \\ \hline
YOLO (L)                                & 0.782          & 0.708          & 0.723  & 0.881          & 0.807          & 0.679          & 0.624          & 0.705        & 0.748          & 0.694          & 0.758          & 0.509          & 0.716                               \\ \hline
\textbf{MAET (w/o ort)}                                & 0.792          & 0.711          & 0.730  & 0.884          & 0.811          & 0.671          & 0.648          & 0.701          & 0.750         & 0.702          & 0.754          & 0.514         & 0.722                               \\ \hline
\textbf{MAET (w ort)}                            & \textbf{0.813}          & \textbf{0.716} & \textbf{0.745}  & \textbf{0.897} & \textbf{0.821} & \textbf{0.695} & 0.655 & \textbf{0.726} & \textbf{0.754} & \textbf{0.727} & \textbf{0.774} & \textbf{0.533} & \textbf{0.740}                      \\ \hline
\end{tabular}
\end{adjustbox}
\caption{Experimental results based on ExDark ~\cite{LOH201930} dataset.  YOLO  (N), YOLO  (L) are the models pretrained using original images/synthetic low-light images and fine-tuned based on the ExDark dataset; MBLLEN \cite{Lv2018MBLLEN}, KIND \cite{kind_kill_the_darkness}, and Zero-DCE \cite{zero_dce}  +  YOLO  (N)  are pre-trained using the original COCO dataset and fine-tuned based on the  Exdark  dataset  processed  by  different  enhancement methods; MAET is our MAET (COCO) finetuned on the Exdark dataset.}\label{tab:EXDark_dataset}
\end{table*}

\subsection{Training Details}

We realize our work based on the open-source object detection toolbox MMDetection ~\cite{mmdetection}. The loss weight components, $\omega_1$ and $\omega_2$, in Eq.~\ref{loss_function} are set to $1$ and $10$, respectively. In this experiment, $L_{obj}$ represents the loss function of YOLO Head output branch in $D_{obj}$, $L_{deg}$ represents the MSE loss of transformation parameters between the prediction of $D_{deg}$ and the known ground truth, as listed in the last line of  Table~\ref{tab:trans_para}: ($k$, $\frac{1}{B}$, $\frac{1}{g_r}$, $\frac{1}{g_b}$, $\frac{1}{\gamma}$), each parameter is normalized in its corresponding category as a pre-process step, and their weights in $L_{deg}$ are set to $5:1:1:1:1$.

All the input images have been cropped and resized to 608 × 608 pixel size. The backbone, DarkNet-53, is initialized with an ImageNet pre-trained model.  We adopt stochastic gradient descent (SGD) as an optimizer and set the image batch size to 8. We set the weight decay to 5e-4 and momentum to 0.9. The learning rate of the encoder ($E$) and object detection decoder ($D_{obj}$) is initially set to 5e-4, and the learning rate of degrading transformation decoder ($D_{deg}$) is initially set to 5e-5. Both the rates adopt a MultiStepLR policy for learning rate decay. For the VOC dataset, we trained our network with a single Nvidia
GeForce RTX 3090 GPU for 50 epochs, and the learning rate decreased by 10 at 20 and 40 epochs, and for the COCO dataset, we trained our network with four 
Nvidia GeForce RTX 3090 GPUs for 273 epochs, and the learning rate decreased by 10 at 218 and 246 epochs.

\subsection{Synthetic Evaluation}
\label{sec:syn_exp}

Pascal VOC ~\cite{voc} is a well-known dataset with 20 categories.
% Following the setting of MMDetection ~\cite{mmdetection}, 
We train our model based on the VOC 2007 and VOC 2012 train and validation sets, and test the model based on the VOC 2007 test set. For VOC evaluation, we report mean average precision (mAP) rate  at IOU threshold of 0.5.

COCO ~\cite{coco_dataset} is another widely used dataset with 80 categories and over 10,0000 images.
% Following ~\cite{mmdetection}, 
We train our model based on the COCO 2017 train set and test the model based on the COCO 2017 validation set. For COCO evaluation, we evaluated each index of COCO dataset. The quantitative results of VOC and COCO datasets are listed in Table~\ref{tab:VOCCOCO}.

In this part, we train and test the YOLOv3 model ~\cite{yolov3} based on the VOC and COCO datasets for normal-lit and synthetic low-lit images as reference. Then, we use the YOLOv3 model trained for normal-lit to test on the set recovered by different low-light enhancement methods ~\cite{Lv2018MBLLEN, kind_kill_the_darkness, zero_dce}\footnote{Here \cite{Lv2018MBLLEN, kind_kill_the_darkness} have been retrained on the normal-lit image and synthetic low-lit image pairs and ~\cite{zero_dce} only trained on low-lit images, because ~\cite{zero_dce} is a self-supervised model which do not need normal-lit ground truth.}. 
 To verify the effectiveness of the orthogonal loss, we train the MAET model with/without orthogonal loss function as MAET (w/o ort) and MAET (w ort) and directly test these models on the low-lit images with no pre-processing.
To ensure fairness, all the methods in the training process are set to same setting parameters, \textit{i.e.},  the data augmentation methods (expand, random crop, multisize, and random flip), input size, learning rate, learning strategy, and training epochs. Experimental configurations and results are listed in Table \ref{tab:VOCCOCO}.

% \usepackage{multirow}

The experimental results in Table \ref{tab:VOCCOCO} show that our MAET has significantly improved the baseline detection framework based on the synthetic low-light dataset. Compared with the enhancement methods~\cite{Lv2018MBLLEN, kind_kill_the_darkness, zero_dce}, the proposed MAET shows superior performance considering all evaluation indexes.  

\subsection{Real-World Evaluation}
\label{sec:rea_world_exp}
To evaluate the performance in a real-world scenario, we have evaluated our trained model (explained in Sec.~\ref{sec:syn_exp}) using the exclusively dark (ExDark) dataset ~\cite{LOH201930}. The dataset includes 7,363 low-light images, ranging from extremely dark environments to twilight with 12 object categories. The local object-bounding boxes are annotated for each image. As EXDark is divided based on different categories, $80\%$ samples of  each  category  are  used  for  fine-tuning on COCO pre-trained model (Sec.\ref{sec:syn_exp}) for 25 epochs with a learning rate of 0.001, and  the  remaining $20\%$ are used for evaluation; we calculate the average precision (AP) of each category (see Table. \ref{tab:EXDark_dataset} for more details) and calculate the overall mean average precision (mAP). Moreover, we have provided some examples in Appendix.A. As listed in Table \ref{tab:EXDark_dataset}, we can see that the proposed MAET method achieves satisfactory performance considering most of the classes and overall mAP. This result affirms that our degrading transformation is in accordance with real-world conditions.

Furthermore, we have evaluated our methods with the $\rm UG^2+$ DARK FACE dataset ~\cite{UG2}; $\rm UG^2+$ is a low-light face detection dataset, which contains 6,000 labeled low-light face images, where 5400 images are used for fine-tuning on the COCO pretrained model (Sec.\ref{sec:syn_exp}) for 20 epochs with a learning rate of 0.001. The other 600 images are used for evaluation; the experiment results are listed in Table \ref{tab:UG}. The proposed MAET method has achieved better results compared with the other methods.

\begin{table}[!htb]
\centering
\begin{tabular}{c|c}
\toprule[1.2pt]
\toprule[1.2pt]
                 & mAP             \\ \hline
YOLO (N)          & 0.483          \\ \hline
MBLLEN ~\cite{Lv2018MBLLEN} + YOLO(N) & 0.516          \\ \hline
KIND ~\cite{kind_kill_the_darkness} + YOLO(N)   & 0.516          \\
\hline
Zero-DCE ~\cite{zero_dce} + YOLO(N)   & 0.542          \\
\hline
YOLO (L)        & 0.540               \\ \hline
\textbf{MAET (w/o ort)}            & 0.542 \\ \hline
\textbf{MAET (w ort)}            & \textbf{0.558} \\ 
\bottomrule[1.2pt]
\bottomrule[1.2pt]
\end{tabular}
\caption{The experiment results on the $\rm UG^2+$ DARK FACE \cite{UG2} dataset.}\label{tab:UG}
\end{table}

\section{Conclusion}

We propose MAET, a novel framework to  explore the intrinsic representation that is equivariant to degradation caused by changes in illumination. The MAET decodes this self-supervised representation to detect objects in a dark environment.  To avoid the over-entanglement of object and degrading features, our method develops a parametric manifold along which multitask predictions can be geometrically formulated by maximizing the orthogonality among the tangents along the output of respective  tasks. Throughout the experiment, the proposed algorithm outperforms the state-of-the-art models pertaining to real-world and synthetic dark image datasets. 

\section*{Acknowledgement}
This work was supported by JST Moonshot R\&D Grant Number JPMJMS2011 and  JST, ACT-X Grant Number
JPMJAX190D, Japan and the National Natural Science Foundation of China under grant 62071333, U1830103, CSTC2018JSCX-MSYBX0115, China

{\small
\bibliographystyle{ieee_fullname}
\bibliography{egpaper_final}

\begin{thebibliography}{10}\itemsep=-1pt

\bibitem{demosaic}
Chapter 5 - comparison of color demosaicing methods.
\newblock In Peter~W. Hawkes, editor, {\em Advances in Imaging and electron
  Physics}, volume 162 of {\em Advances in Imaging and Electron Physics}, pages
  173--265. Elsevier, 2010.

\bibitem{ANAYA2018NENOIR}
Josue Anaya and Adrian Barbu.
\newblock Renoir – a dataset for real low-light image noise reduction.
\newblock {\em Journal of Visual Communication and Image Representation},
  51:144 -- 154, 2018.

\bibitem{brooks2019unprocessing}
Tim Brooks, Ben Mildenhall, Tianfan Xue, Jiawen Chen, Dillon Sharlet, and
  Jonathan~T Barron.
\newblock Unprocessing images for learned raw denoising.
\newblock In {\em Proceedings of the IEEE Conference on Computer Vision and
  Pattern Recognition}, pages 11036--11045, 2019.

\bibitem{see_in_the_dark}
C. {Chen}, Q. {Chen}, J. {Xu}, and V. {Koltun}.
\newblock Learning to see in the dark.
\newblock In {\em 2018 IEEE/CVF Conference on Computer Vision and Pattern
  Recognition}, pages 3291--3300, 2018.

\bibitem{mmdetection}
Kai Chen, Jiaqi Wang, and et.al.
\newblock {MMDetection}: Open mmlab detection toolbox and benchmark.
\newblock {\em arXiv preprint arXiv:1906.07155}, 2019.

\bibitem{voc}
Mark Everingham, Luc Gool, Christopher~K. Williams, John Winn, and Andrew
  Zisserman.
\newblock The pascal visual object classes ({VOC}) challenge.
\newblock {\em Int. J. Comput. Vision}, 88(2):303–338, June 2010.

\bibitem{Noise_model}
A. {Foi}, M. {Trimeche}, V. {Katkovnik}, and K. {Egiazarian}.
\newblock Practical {P}oissonian-gaussian noise modeling and fitting for
  single-image raw-data.
\newblock {\em IEEE Transactions on Image Processing}, 17(10):1737--1754, 2008.

\bibitem{zero_dce}
C. {Guo}, C. {Li}, J. {Guo}, C.~C. {Loy}, J. {Hou}, S. {Kwong}, and R. {Cong}.
\newblock Zero-reference deep curve estimation for low-light image enhancement.
\newblock In {\em 2020 IEEE/CVF Conference on Computer Vision and Pattern
  Recognition (CVPR)}, pages 1777--1786, 2020.

\bibitem{LIME_Illumination_Map_Estimation}
X. {Guo}, Y. {Li}, and H. {Ling}.
\newblock Lime: Low-light image enhancement via illumination map estimation.
\newblock {\em IEEE Transactions on Image Processing}, 26(2):982--993, 2017.

\bibitem{ISP_flexisp}
Felix Heide and Steinberger et.al.
\newblock Flex{ISP}: A flexible camera image processing framework.
\newblock {\em ACM Trans. Graph.}, 33(6), Nov. 2014.

\bibitem{ISP_L3method}
H. {Jiang}, Q. {Tian}, J. {Farrell}, and B.~A. {Wandell}.
\newblock Learning the image processing pipeline.
\newblock {\em IEEE Transactions on Image Processing}, 26(10):5032--5042, 2017.

\bibitem{enlightengan}
Yifan Jiang, Xinyu Gong, Ding Liu, Yu Cheng, Chen Fang, Xiaohui Shen, Jianchao
  Yang, Pan Zhou, and Zhangyang Wang.
\newblock Enlighten{GAN}: Deep light enhancement without paired supervision.
\newblock {\em IEEE Transactions on Image Processing}, 30:2340--2349, 2021.

\bibitem{multiscale_retinex}
D.~J. {Jobson}, Z. {Rahman}, and G.~A. {Woodell}.
\newblock A multiscale retinex for bridging the gap between color images and
  the human observation of scenes.
\newblock {\em IEEE Transactions on Image Processing}, 6(7):965--976, 1997.

\bibitem{ISP_ECCV_2016_brown}
Hakki~Can Karaimer and Michael~S. Brown.
\newblock A software platform for manipulating the camera imaging pipeline.
\newblock In {\em European Conference on Computer Vision (ECCV)}, 2016.

\bibitem{low_light_gan}
G. {Kim}, D. {Kwon}, and J. {Kwon}.
\newblock Low-light{GAN}: Low-light enhancement via advanced generative
  adversarial network with task-driven training.
\newblock In {\em 2019 IEEE International Conference on Image Processing
  (ICIP)}, pages 2811--2815, 2019.

\bibitem{krizhevsky2017imagenet}
Alex Krizhevsky, Ilya Sutskever, and Geoffrey~E Hinton.
\newblock Imagenet classification with deep convolutional neural networks.
\newblock {\em Communications of the ACM}, 60(6):84--90, 2017.

\bibitem{VJ_detector_low_light}
Roman. {Kvyetnyy}, Roman. {Maslii}, and et.al.
\newblock Object detection in images with low light condition.
\newblock In Ryszard~S. Romaniuk and Maciej Linczuk, editors, {\em Photonics
  Applications in Astronomy, Communications, Industry, and High Energy Physics
  Experiments 2017}, volume 10445, pages 250 -- 259. International Society for
  Optics and Photonics, SPIE, 2017.

\bibitem{retinex}
Edwin~H. Land.
\newblock An alternative technique for the computation of the designator in the
  retinex theory of color vision.
\newblock {\em Proceedings of the National Academy of Sciences of the United
  States of America}, 83(10):3078--3080, 1986.

\bibitem{local_HE}
C. {Lee}, C. {Lee}, and C. {Kim}.
\newblock Contrast enhancement based on layered difference representation of
  {2D} histograms.
\newblock {\em IEEE Transactions on Image Processing}, 22(12):5372--5384, 2013.

\bibitem{LightenNet}
Chongyi Li, Jichang Guo, Fatih Porikli, and Yanwei Pang.
\newblock Lightennet: A convolutional neural network for weakly illuminated
  image enhancement.
\newblock {\em Pattern Recognition Letters}, 104:15--22, 2018.

\bibitem{mobile_camera_imaging2}
Orly Liba, Kiran Murthy, Yun-Ta Tsai, Tim Brooks, Tianfan Xue, Nikhil Karnad,
  Qiurui He, Jonathan~T. Barron, Dillon Sharlet, Ryan Geiss, Samuel~W.
  Hasinoff, Yael Pritch, and Marc Levoy.
\newblock Handheld mobile photography in very low light.
\newblock {\em ACM Trans. Graph.}, 38(6), Nov. 2019.

\bibitem{retinanet}
T. {Lin}, P. {Goyal}, R. {Girshick}, K. {He}, and P. {Dollár}.
\newblock Focal loss for dense object detection.
\newblock In {\em 2017 IEEE International Conference on Computer Vision
  (ICCV)}, pages 2999--3007, 2017.

\bibitem{coco_dataset}
Tsung-Yi Lin, Michael Maire, Serge Belongie, James Hays, Pietro Perona, Deva
  Ramanan, Piotr Doll{\'a}r, and C.~Lawrence Zitnick.
\newblock Microsoft {COCO}: Common objects in context.
\newblock In David Fleet, Tomas Pajdla, Bernt Schiele, and Tinne Tuytelaars,
  editors, {\em Computer Vision -- ECCV 2014}, pages 740--755, Cham, 2014.
  Springer International Publishing.

\bibitem{xiumei}
Yajing Liu, Xinmei Tian, Ya Li, Zhiwei Xiong, and Feng Wu.
\newblock Compact feature learning for multi-domain image classification.
\newblock In {\em 2019 IEEE/CVF Conference on Computer Vision and Pattern
  Recognition (CVPR)}, pages 7186--7194, 2019.

\bibitem{LOH201930}
Yuen~Peng Loh and Chee~Seng Chan.
\newblock Getting to know low-light images with the exclusively dark dataset.
\newblock {\em Computer Vision and Image Understanding}, 178:30 -- 42, 2019.

\bibitem{LLNet}
Kin~Gwn Lore, Adedotun Akintayo, and Soumik Sarkar.
\newblock Llnet: A deep autoencoder approach to natural low-light image
  enhancement.
\newblock {\em Pattern Recognition}, 61:650 -- 662, 2017.

\bibitem{lv2021attention}
Feifan Lv, Yu Li, and Feng Lu.
\newblock Attention guided low-light image enhancement with a large scale
  low-light simulation dataset.
\newblock {\em International Journal of Computer Vision}, 129(7):2175--2193,
  2021.

\bibitem{Lv2018MBLLEN}
Feifan Lv, Feng Lu, Jianhua Wu, and Chongsoon Lim.
\newblock {MBLLEN}: Low-light image/video enhancement using {CNN}s.
\newblock In {\em British Machine Vision Conference (BMVC)}, 2018.

\bibitem{traditional_feature_method}
Yasushi Makihara, Masao Takizawa, Yoshiaki Shirai, and Nobutaka Shimada.
\newblock Object recognition under various lighting conditions.
\newblock In Josef Bigun and Tomas Gustavsson, editors, {\em Image Analysis},
  pages 899--906, Berlin, Heidelberg, 2003. Springer Berlin Heidelberg.

\bibitem{Unconstrained_Face_Detection}
H. {Nada}, V.~A. {Sindagi}, H. {Zhang}, and V.~M. {Patel}.
\newblock Pushing the limits of unconstrained face detection: a challenge
  dataset and baseline results.
\newblock In {\em 2018 IEEE 9th International Conference on Biometrics Theory,
  Applications and Systems (BTAS)}, pages 1--10, 2018.

\bibitem{NightOwls}
Lukas Neumann, Michelle Karg, Shanshan Zhang, Christian Scharfenberger, Eric
  Piegert, Sarah Mistr, Olga Prokofyeva, Robert Thiel, Andrea Vedaldi, Andrew
  Zisserman, and Bernt Schiele.
\newblock Night{O}wls: A pedestrians at night dataset.
\newblock In {\em Asian Conference on Computer Vision}, 2018.

\bibitem{isp_icip_2018}
J. {Nishimura}, T. {Gerasimow}, R. {Sushma}, A. {Sutic}, C. {Wu}, and G.
  {Michael}.
\newblock Automatic {ISP} image quality tuning using nonlinear optimization.
\newblock In {\em 2018 25th IEEE International Conference on Image Processing
  (ICIP)}, pages 2471--2475, 2018.

\bibitem{jigsaw_selfsupervised}
Mehdi Noroozi and Paolo Favaro.
\newblock Unsupervised learning of visual representations by solving jigsaw
  puzzles.
\newblock In {\em European Conference on Computer Vision}, pages 69--84.
  Springer, 2016.

\bibitem{pathak2016context}
Deepak Pathak, Philipp Krahenbuhl, Jeff Donahue, Trevor Darrell, and Alexei~A
  Efros.
\newblock Context encoders: Feature learning by inpainting.
\newblock In {\em Proceedings of the IEEE conference on computer vision and
  pattern recognition}, pages 2536--2544, 2016.

\bibitem{Plotz17CVPR}
T. Plotz and S. Roth.
\newblock Benchmarking denoising algorithms with real photographs.
\newblock In {\em 2017 IEEE Conference on Computer Vision and Pattern
  Recognition (CVPR)}, pages 2750--2759, Los Alamitos, CA, USA, jul 2017. IEEE
  Computer Society.

\bibitem{poynton2003digital}
C. Poynton, Inc Books24x7, and Engineering~Information Inc.
\newblock {\em Digital Video and HD: Algorithms and Interfaces}.
\newblock Computer Graphics. Elsevier Science, 2003.

\bibitem{qi2019avt}
Guo-Jun Qi, Liheng Zhang, Chang~Wen Chen, and Qi Tian.
\newblock {AVT}: Unsupervised learning of transformation equivariant
  representations by autoencoding variational transformations.
\newblock In {\em Proceedings of the IEEE International Conference on Computer
  Vision}, pages 8130--8139, 2019.

\bibitem{ISP_2005_magazine}
R. {Ramanath}, W.~E. {Snyder}, Y. {Yoo}, and M.~S. {Drew}.
\newblock Color image processing pipeline.
\newblock {\em IEEE Signal Processing Magazine}, 22(1):34--43, 2005.

\bibitem{yolov3}
Joseph Redmon and Ali Farhadi.
\newblock {YOLO}v3: An incremental improvement.
\newblock {\em CoRR}, abs/1804.02767, 2018.

\bibitem{fasterrcnn}
Shaoqing Ren, Kaiming He, Ross Girshick, and Jian Sun.
\newblock Faster {R-CNN}: Towards real-time object detection with region
  proposal networks.
\newblock In {\em Proceedings of the 28th International Conference on Neural
  Information Processing Systems - Volume 1}, NIPS'15, page 91–99, Cambridge,
  MA, USA, 2015. MIT Press.

\bibitem{yolo-in-the-dark}
Yukihiro Sasagawa and Hajime Nagahara.
\newblock Yolo in the dark - domain adaptation method for merging multiple
  models -.
\newblock In {\em Proceedings - European Conference on Computer Vision}, August
  2020.

\bibitem{shen2017msrnetlowlight}
Liang Shen, Zihan Yue, Fan Feng, Quan Chen, Shihao Liu, and Jie Ma.
\newblock Msr-net:low-light image enhancement using deep convolutional network,
  2017.

\bibitem{suteu2019regularizing}
Mihai Suteu and Yike Guo.
\newblock Regularizing deep multi-task networks using orthogonal gradients.
\newblock {\em arXiv preprint arXiv:1912.06844}, 2019.

\bibitem{Global_HE}
C. {Tomasi} and R. {Manduchi}.
\newblock Bilateral filtering for gray and color images.
\newblock In {\em Sixth International Conference on Computer Vision (IEEE Cat.
  No.98CH36271)}, pages 839--846, 1998.

\bibitem{wang2019enaet}
Xiao Wang, Daisuke Kihara, Jiebo Luo, and Guo-Jun Qi.
\newblock En{AET}: Self-trained ensemble autoencoding transformations for
  semi-supervised learning.
\newblock {\em arXiv preprint arXiv:1911.09265}, 2019.

\bibitem{Chen2018Retinex}
Chen Wei, Wenjing Wang, Wenhan Yang, and Jiaying Liu.
\newblock Deep retinex decomposition for low-light enhancement.
\newblock In {\em British Machine Vision Conference}, 2018.

\bibitem{low_light_saliency}
Xin Xu, Shiqin Wang, Zheng Wang, Xiaolong Zhang, and Ruimin Hu.
\newblock Exploring image enhancement for salient object detection in low light
  images.
\newblock {\em arXiv preprint arXiv:2007.16124}, 2020.

\bibitem{UG2}
W. {Yang}, Y. {Yuan}, and et.al.
\newblock Advancing image understanding in poor visibility environments: A
  collective benchmark study.
\newblock {\em IEEE Transactions on Image Processing}, 29:5737--5752, 2020.

\bibitem{yu2020gradient}
Tianhe Yu, Saurabh Kumar, Abhishek Gupta, Sergey Levine, Karol Hausman, and
  Chelsea Finn.
\newblock Gradient surgery for multi-task learning.
\newblock {\em arXiv preprint arXiv:2001.06782}, 2020.

\bibitem{zhang2019aet}
Liheng Zhang, Guo-Jun Qi, Liqiang Wang, and Jiebo Luo.
\newblock {AET} vs. {AED}: Unsupervised representation learning by
  auto-encoding transformations rather than data.
\newblock In {\em Proceedings of the IEEE Conference on Computer Vision and
  Pattern Recognition}, pages 2547--2555, 2019.

\bibitem{kind_kill_the_darkness}
Yonghua Zhang, Jiawan Zhang, and Xiaojie Guo.
\newblock Kindling the darkness: A practical low-light image enhancer.
\newblock MM '19, New York, NY, USA, 2019. Association for Computing Machinery.

\bibitem{optical_flow_in_the_dark}
Y. {Zheng}, M. {Zhang}, and F. {Lu}.
\newblock Optical flow in the dark.
\newblock In {\em 2020 IEEE/CVF Conference on Computer Vision and Pattern
  Recognition (CVPR)}, pages 6748--6756, 2020.

\bibitem{zheng_2020_ECCV}
Ziqiang Zheng, Yang Wu, Xinran Han, and Jianbo Shi.
\newblock Fork{GAN}: Seeing into the rainy night.
\newblock In {\em The IEEE European Conference on Computer Vision (ECCV)},
  August 2020.

\bibitem{centernet}
Xingyi Zhou, Dequan Wang, and Philipp Kr{\"a}henb{\"u}hl.
\newblock Objects as points.
\newblock In {\em arXiv preprint arXiv:1904.07850}, 2019.

\end{thebibliography}
}

\clearpage

\section{Display of Object Detection Results in Real-World Situation}
\label{sec:display}

\begin{figure}[b]
    \centering
    \includegraphics[width = 8.5cm, height = 6.5cm]{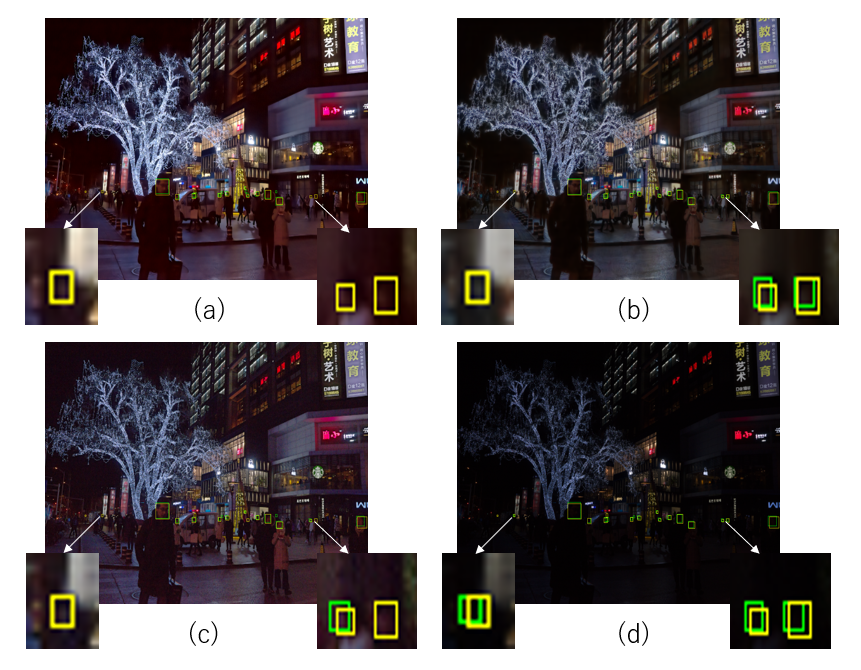}
    \caption{Detection results of $\rm UG^2+$ DARK FACE dataset ~\cite{UG2}. (a)/(b)/(c) is the detection result of YOLO based on the dataset pre-processed by MBLLEN ~\cite{Lv2018MBLLEN}/KIND ~\cite{kind_kill_the_darkness}/Zero-DCE \cite{zero_dce} and (d) is the is the detection result of MAET-YOLO model on original dataset, yellow and green boxes are ground truth boxes and prediction results, respectively.}
    \label{fig:Ug2Dark-FAce}
    % \setlength{\belowcaptionskip}{-2cm}
    % \vspace{-0.5cm}
\end{figure}

\begin{figure*}[]
    \centering
    \includegraphics[width = 16.5cm, height = 12.5cm]{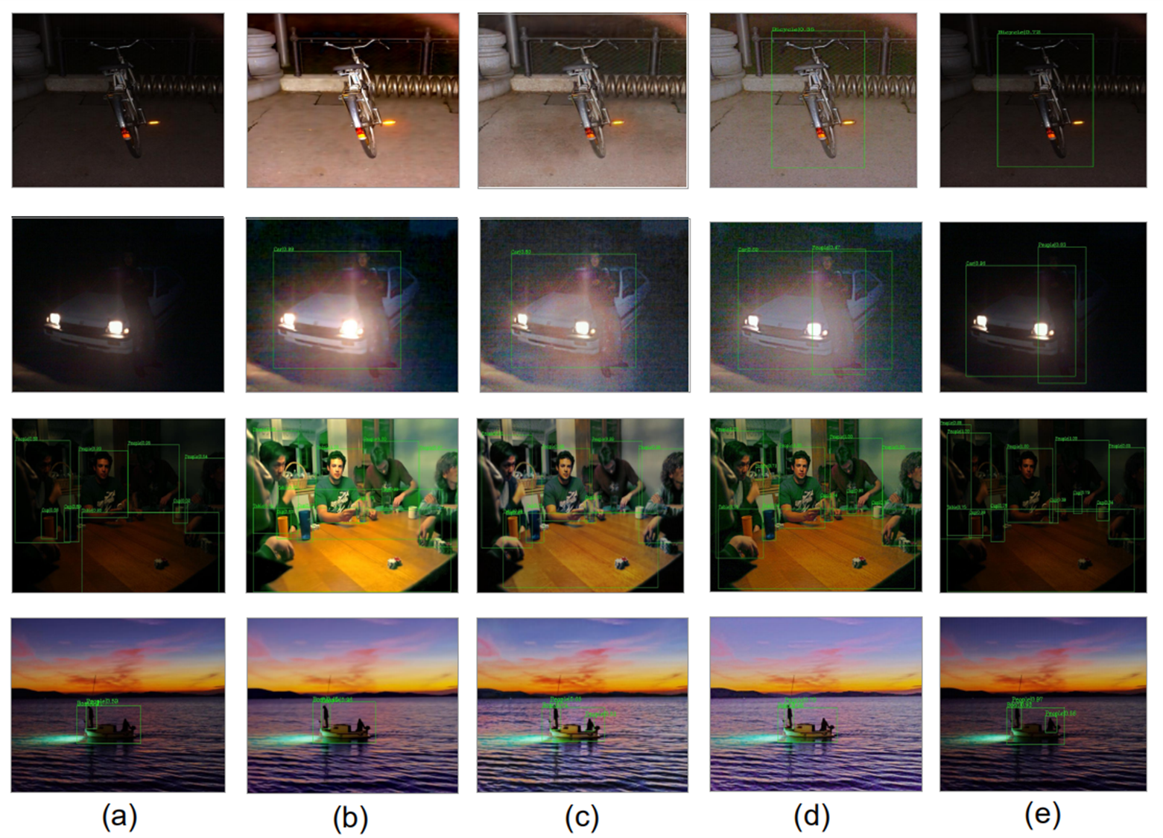}
    \caption{Detection results of ExDark ~\cite{LOH201930} dataset: (a) is the detection result of YOLO using the original ExDark dataset, (b)/(c)/(d) is the detection result of YOLO based on the ExDark dataset pre-processed by MBLLEN ~\cite{Lv2018MBLLEN}/KIND ~\cite{kind_kill_the_darkness}/Zero-DCE ~\cite{zero_dce}, and (e) is the detection result of the proposed MAET–YOLO method based on the original ExDark dataset.}
    \label{fig:exdark}
\end{figure*}

In this section we would give more examples on real data sets. Fig.\ref{fig:Ug2Dark-FAce} shows a case in $\rm UG^2+$ DARK FACE ~\cite{UG2} dataset. Fig.\ref{fig:exdark} shows the detection results of ExDark ~\cite{LOH201930} dataset. Our MAET has shown better detection performance when facing  actual dark light detection tasks.

% \vspace{-0.8cm}

\section{Ablation Study}
\subsection{Other Low-Light Image Synthesis Methods}
\label{sec:data_synthesis}

Although the purpose of this paper is not for a  perfect low-light image synthesis pipeline, we have also compared with existing  low light image synthesis methods (from normal light sRGB to low light sRGB) and evaluate their impact  on low-light object detection tasks.

The work in ~\cite{LightenNet} proposed  to use the RetiNex model ~\cite{retinex} to generate low light counterpart by normal light images:
\begin{equation}
    I(x) = R(x) \cdot L,
\end{equation}
in this equation, $R(x)$ is the clear normal-lit images (same as $x$ in Eq.15), $I(x)$ is the generated low-lit counterpart (same as $t_{deg}(x)$ in Eq.15) and $L$ is the random fixed illumination value, here $L$, same as our parameter $k$'s range.

The work in ~\cite{LLNet, Lv2018MBLLEN, optical_flow_in_the_dark} proposed  to use an invert gamma correction with additional noises to generate low light degraded image from normal light counterpart, the equation shown as follow:
\begin{equation}
    t_{deg}(x) = x^{\gamma} + n.
\end{equation}

\begin{equation}
    t_{deg}(x) = x \cdot k.
\end{equation}
here $\gamma$ is the gamma curve parameter 
% ($\gamma \sim U(1.8, 2.5)$) 
and n is the additional noise (Poisson noise or Gaussian-Poisson noise model).

\subsection{Demosacing's Influence}
\label{sec:mosaic}
Demosacing is an essential part in camera image signal processing pipeline ~\cite{demosaic, ISP_ECCV_2016_brown, brooks2019unprocessing}, which aims to recover intermediate gray-scale image to the R/G/B value by interpolating the missing values in Bayer pattern. Unlike the previous work ~\cite{brooks2019unprocessing}, we ignore this step for simplicity in our ISP procedure. In supplementary material, we show an example after adding mosaicing process after invert WB process ((d) in Fig.3) and demosacing process after WB process ((f) in Fig.3).

%Different like the work in \cite{brooks2019unprocessing}, we don't take mosaicing and demosaicing in our low light data generation pipeline. This is because in actual operation, mosaicing part would reduce the images' resolution and interpolation process in demosaicing part would reduce the images' sharpness and quality, this would affect the back-end object detection performance, following \cite{ISP_ECCV_2016_brown} and \cite{brooks2019unprocessing}, we add mosaicing process after invert WB process ((e) in Fig.\ref{fig:pipeline}) and demosacing process after WB process ((g) in Fig.\ref{fig:pipeline}),
%some examples have been shown in . \red{Here to add a figure, show that mosacing and demosacing's effects on resolution and sharpness~}

To evaluate effects of different low light data generation methods: Retinex based generation method $L_{R}$, inverse gamma curve $L_{G}$, inverse gamma curve with additional possion noise $L_{GP}$, inverse gamma curve with additional mixed Gaussian-Poisson noise model $L_{Gmix}$, our proposed data generation method in Sec.3.3 $L_{ours}$, and our proposed data generation method with mosaicing and demosacing process $L_{ours+m}$. We measured the performance of using different dark light data on the real world datasets ~\cite{LOH201930, UG2}, shown as YOLO (L) in Table \ref{tab:low_light_data}, the training configurations and strategies are same as Sec.4, it could be seen that our synthetic method is of greatest help in improving the detection performance of real datasets \cite{LOH201930}, \cite{UG2}.

% \usepackage{multirow}
\begin{table}[]
\centering
\begin{adjustbox}{max width= 1 \linewidth}
\begin{tabular}{c|c|cc}
\toprule[1.2pt]
\toprule[1.2pt]
                            &        & ExDark & DARK FACE               \\ \hline
\multirow{6}{*}{YOLO (L)} & $L_R$            &      0.698   &       0.511           \\ \cline{2-4} 
                            & $L_G$         &      0.709   &       0.528           \\ \cline{2-4} 
                            & $L_{GP}$          &     0.712      &      0.532            \\ \cline{2-4} 
                            & $L_{Gmix}$     &       0.713  &        0.535          \\ \cline{2-4} 
                            & \bm{$L_{ours+m}$} &      0.706           &        0.530          \\ \cline{2-4} 
                            & \bm{$L_{ours}$}     &   \textbf{0.716}  &      \textbf{0.540}           \\ 
\bottomrule[1.2pt]
\bottomrule[1.2pt]
\end{tabular}
\end{adjustbox}
\vspace{+0.3cm}
\caption{The experiment results on ExDark ~\cite{LOH201930} dataset and $\rm UG^{2+}$  DARK FACE \cite{UG2} datasets by using different kinds of synthetic low light COCO \cite{coco_dataset} dataset. The detection results verify the reliability of our synthesis method. }\label{tab:low_light_data}
\end{table}

\end{document}